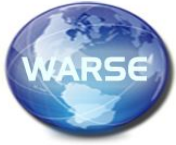

# A Study on various state of the art of the Art Face Recognition System using Deep Learning Techniques

**Sukhada Chokkadi[1], Sannidhan MS[2], Sudeepa K B[3], Abhir Bhandary[4]**
[1]NMAMIT, India, sukhada.chokkadi@gmail.com
[2]NMAMIT, India, sannidhan@nitte.edu.in
[3]NMAMIT, India, sudeepa@nitte.edu.in
[4] NMAMIT, India, abhirbhandary@nitte.edu.in

**ABSTRACT**

Considering the existence of very large amount of available data repositories and reach to the very advanced system of hardware, systems meant for facial identification have evolved enormously over the past few decades. Sketch recognition is one of the most important areas that have evolved as an integral component adopted by the agencies of law administration in current trends of forensic science. Matching of derived sketches to photo images of face is also a difficult assignment as the considered sketches are produced upon the verbal explanation depicted by the eye witness of the crime scene and may have scarcity of sensitive elements that exist in the photograph as one can accurately depict due to the natural human error. Substantial amount of the novel research work carried out in this area up late used recognition system through traditional extraction and classification models. But very recently, few researches work focused on using deep learning techniques to take an advantage of learning models for the feature extraction and classification to rule out potential domain challenges. The first part of this review paper basically focuses on deep learning techniques used in face recognition and matching which as improved the accuracy of face recognition technique with training of huge sets of data. This paper also includes a survey on different techniques used to match composite sketches to human images which includes component-based representation approach, automatic composite sketch recognition technique etc.

**Key words:** Face recognition, Convolution Neural Networks (CNN), Sketch recognition, composite sketches, deep learning

## 1.INTRODUCTION

As per the researches carried out, a complete face recognition system includes two patterns of face detection and face recognition: 1) Structural similarity and 2) individual local differences of human faces. Therefore, it is required to extract the features of the face through the face detection process. The evolution of face recognition is due to its technical challenges and huge potential application in video surveillance, identity authorization, multimedia applications, home and office security, law enforcement and different human-computer interaction activities. Facial recognition technology (FRT) is one of the most controversial new tools. It was first developed in the 1960s. It has recently become accessible to the mass market—to both law enforcement and private consumers. FRT has the capacity to eliminate the need for passwords, fingerprint data, and even keys. However, it could also end privacy. Face recognition technology has also developed research and implementation based on mobile phone system. [1]

Surveys carried out have proved that the United States suffers from a systemic problem of racial disparities in traffic stops, stop and frisks, and arrests. Due to this African Americans are arrested at twice the rate and shrivelled at nearly three times the rate of any other race. Face recognition is one of the few biometric methods that possess the merits of both high accuracy and low intrusiveness. Biometric based time attendance system has been developed using face recognition technology [2-3]. Hence the demand for automatic face recognition system is increasing potentially in every single area. Automatic face recognition involves: 1) face detection, 2) feature extraction and 3) face recognition. Face recognition algorithms are broadly classified into two classes: 1) template based system 2) based on geometric feature. Template dependent procedures calculate value of the correlation among the face and more than one model templates to match the facial identity. Principal Component Analysis (PCA), Linear Discriminate Analysis (LDA), methods of kernel based
etc. were applied to compose the templates corresponding to the face. Face recognition is an initial stage in the identification process [5-6].

Sketch recognition system up late evolved to be the next level of facial recognition system in forensic sciences and law enforcement agencies in solving crime related issues. Sketches are basically classified into three different types: 1) Viewed sketch that is generated by an artist looking to the subjects face or photograph. 2) Forensic sketch is another type, that is generated by the trained artist based on the descriptions given by the witness of the crime scene 3) Composite sketches are the computer-generated sketches which are generated by using software programs. In the current scenario, computer system produced sketches are being chosen since they proved to be faster to construct and also accurate when compared to the traditional hand drawn sketches. Researchers have again proved that, matching computer generated sketches to facial photo images is also a difficult work as these sketches being generated and corresponding photos belongs to two different domains and also composite sketches may lack many of the tiny descriptions that are actually found in the photograph. Research paper [4] gives description about technique used in matching composite images to human faces using transfer learning. With the advent of this technology, these techniques algorithms are utilized in several applications and well-being agendas as well as law administration systems like the security forces of the border and other systems related to forensic science.

Face recognition techniques have taken a drift undoubtedly over the years. Traditional methods were dependent on features, like descriptors, combined with machine learning techniques, such as principal component analysis (PCA), support vector machines SVM) etc. Huge variations in facial features motivated research community to implement some specific systems to handle the variations. Specially designed systems such as time invariant systems, variation in pose factors and variation in illumination factors etc. were made use. Currently, models developed for face recognition systems are being reinstated and influenced by Deep Learning (DL) methods based on CNNs (Convolutional Neural Networks) [7-10]. The huge plus point of using these methodologies is that they are learnable through training process of huge datasets simply reachable over the internet. Apart from this efficiency and performance of the developed systems using Deep Learning concepts is relatively higher than the methods implemented using traditional techniques[11-14]. Additionally, CNNs can also solve broader range of challenging computer vision problems. Considering the a fore mentioned facts, in this paper we have conducted a comprehensive review on the Deep Learning techniques that are of late used in the field of facial recognition system under number of applications and number of domains [15-16].

## 2 A COMPREHENSIVE REVIEW ON VARIOUS FACE RECOGNITION SYSTEM USING DEEP LEARNING

In this paper, on conducting an exhaustive literature review, we have identified some best and also latest techniques of Deep Learning particular specifically used in the research articles of Facial Recognition System which are suited for a type of facial recognition system working better to suffice the requirement of the problem having their own limitations under certain scenarios.

In their research paper [17], authors have proposed a technique called joint fine lining in Deep Neural Networks (DNN) used for the system of facial recognition. As per the proposed methodology, technique uses two levels of deep networks. One of the deep networks captures temporary features related to the appearances from the sequences of images were as another one derives geometrical traits. As a result of this, researchers were successful in developing a new integration method for facial recognition. Proposed integration system proved better accuracy by yielding better results than the traditional methods implemented prior to this achieving a result of 97.25% accuracy.

Authors in their article [18], proposed a type of Deep Learning technique for facial recognition called as "regularised deep learning" used for body weight variations which is a natural and intrinsic feature of an individual's process of aging. In this research article, a regularised dependent approach is used to study facial representations corresponding to the variation in weight using two deep learning architectures of different types. 1) A Sparse stacked demonizing auto encoder are applied to carry out the training of a complete deep network. Later Stacking is achieved in an efficient way akin that the generated output of the initial encoder is provided as an input to the next one 2) Deep Boltzmann machines is used in unsupervised learning for feature representation from the given set. The paper has very efficiently addressed the challenge of obtaining large labelled data for the purpose of training for variations in body weight using a regularised deep learning. The proposed framework with accuracy of 26.0% and of 65.8% for Rank1 and Ran10 respectively.

In an article [19], authors have presented a technique on how the rotating of face is done utilizing a Multi-task Deep Neural Network (M-DNN). This research article presents a innovative system of architecture dependent on multitask system of learning that can earn very large performance in carrying out the rotation of a face image of a target pose. The experiment section consists of 4 parts. In the first part, feature space of each layer is to analyze and study the input. Then, the target image is constructed to understand and preserve the identity. In the third stage, multi-model is compared to a single model. Finally, experiment is conducted to show the advantages. The input image is reconstructed after the completion of first task model rotates an image to a particular pose. In the recognition task in connection to the random poses and the factors of illuminations, the model presented provably wins against any of the previous traditional techniques more than 4~6%.

In their research article [20], authors have proposed a system for robust facial recognition technique under varying illumination conditions. In this research article, authors have concentrated on the problems concerning to the textural based illumination handling for the face recognition under both indoor and outdoor lighting conditions. To address the problem, this, authors have implemented a technique that can handle the noise produced with highest recognition rate. A standard Histogram equalization technique was adopted to remove illumination from the input facial images. The method is most popular datasets of faces. In the dataset used, some of the front views under changing states were used for evaluation. To extract the features, Linear Discriminant Analysis (LDA) and Kernel Discriminant Analysis (KDA) were made use to exhibit the front views in low quality. The rate of recognition on each of the database represent that the version corresponding to a kernel of LDA gives comparatively more results than plain LDA. The result achieved is 7 to 8% high. This particular method achieves an accuracy of about 93% when compared with all the existing methods.

In their article [21], authors have proposed a Facial Expression Recognition via Deep Learning concept. Proposed technique of deep learning is implemented using a Convolution Neural Network (CNN) which is one of the most effective techniques in the field of deep learning. Proposed methodology works under two different steps once after preparing the datasets required for the experiment: 1) In the first step, batch size for the CNN is fixed in terms of the number of input sets supplied. 2) In the second step, the technique concentrated on improving the classification of the models. As a part of improvement, proposed technique focused on repetition of training of CNN architecture to fine tune the initial model. This method accomplishes the best results on CK+ database with an accuracy of 71.04%.

In their research article [22], authors have presented a comprehensive analysis of Deep Learning based Neural Network (CNN) architecture known as Visual Geometry Group (VGG) Face Network is used for the face recognition. VGG Face network consists of 2.6 million trained facial images. Results presented in the paper proved the achievement of 98.95% accuracy.

In an article [23], authors have proposed the technique of the unconstrained face verification using Deep Convolution Neural Network (DCNN) features. Based on deep convolution features, algorithm for corrupt face authentication is implemented in the paper. This method consists of two facial features. Training and Testing. As per the proposed methodology, initially, face and landmark detection is performed and then, training of DCNN is performed. Given a pair of images, match score based on the closeness of the DCNN features is computed. Processing, Deep face feature representation followed by Joint Bayesian Metric Learning is used in obtaining the final results. The DCNN model for some days, the DCNN model was trained using NVidia Tesla K40. To extract the various features, it took about 0.006 second for every facial image with 97.15% mean accuracy.

Authors in their research article [24] proposed a technique based on deep learning for the changing display and format of each facial unit. In this article, they have implemented a novel approach for the detection of facial acting unit using a combination of convolution and bidirectional memory neural networks. The suggested approach used training of minute regions of the image and similar images of binary version to study the shape and appearance. To represent the changing behaviour, authors used an array of consecutive images as input to Convolutional Neural

Network (CNN), an altered array of image sectors and masks representing binary type. The features studied from this CNN are further used for the training Bi-directional Long Short-Term Memory (BLSTM) Neural Networks. The decision value is obtained from the BLSTM network. Performance from the proposed method was relatively greater on the SEMAINE dataset.

In their research paper [25], researchers have proposed a system that aims in knowing a type of deep learning called multi-instance deep learning where regional divisions were discovered for the identification of body part. This frame work aims at 1] discovering the neighbourhood locations 2] learning an image level classifier based on these local regions. The first part of the research in the article aims at learning the classic regional characteristics in a supervised learning style. In the next part of the research article, some selective, information scarce neighbourhood images were captured from the images. Finally the run time image categorization was understood adopting a trained Convolution Neural Network (CNN) model. BCNN2 attains the best results among all the other methods with 99.9% recall and precision.

In their research article [26], authors have implemented a type of feature being driven using the concept of Deep Neural Network (DNN), basically learning method for various views of the facial expression. In the proposed technique, Scale Invariant Feature Transform (SIFT) characteristics, analogous to a fixed number of notable regions of every face photo were initially captured from every input considered. The proposed model also employs multiple layers to character which maps the correlation between the semantic information of the SIFT vectors and the SIFT vectors themselves. Experimental results are extracted out of BU-3DFE and multi-pie facial recognition database containing various facial expressions. The accuracy may be different with different methods. However, if the augmented samples are used, this deep neural network can attain the precision of about 85.2%. This is the highest among these methods.

Yu, H., Luo, Z., & Tang, Y., i their research article [27], proposed a technique for a face recognition task, there are very few resources used to learn Deep data face model. There are very limited training samples for a face recognition task. These problems were addressed through transferring an already learned model of the facial image using deep learning system considered as a initial type of model. Then, higher layer depictions were learned on a minute and a distinct training set. The aimed target model is obtained. The target model achieved 0% error rates on all the three datasets that were used, with very less training samples for every person's image.

In an article [28], authors have identified the problems existing in the area of forged face images and the aspects that may lead to the variations during the face recognition like noise, deviations in the angle etc. are the primary reasons for the system to lose its transcendence. In this particular method, deep learning was introduced in the sense of providing thorough study about the face patterns existing in the system. The training database matrix and the facial images consisting of columns and rows are present. The input images can be reconstructed. Errors or alterations arising because of the reorganization in each of the classes in specifically noted. Further, the data projection matrix is determined by gauging every face image to its corresponding substitute space that is learned outlearned subspace. The proposed method accepts to evaluate every sample existing in the system. Further, it keeps the record of data from the facial characteristics and performs classification. It was observed that 92.8% accuracy was gained on YALE face datasets by implementing the suggested algorithm.

In their research article [29], authors have presented an active face recognition model using Deep Learning based linear discriminant classification (LDC). Proposed system was implemented using a matrix formed out of the database that corresponds to facial system comprising of columns and rows. It also enhanced the accuracy of the LDRC by keeping a record of the previously trained faces. The suggested method produced improved recognition and separation of biometric face images than the other algorithms built earlier. 92.8% accuracy was accomplished on YALE face datasets by implementing this algorithm.

In an article [30], authors in their research work has proposed a system for automatic facial sketch to photo matching system. Entire work in this article was divided into two broad steps: In the first step, set of facial sketches are constructed. Later, in the second step, an automated facial sketch to facial photo corresponding system was built. The proposed article also compares the accuracy between matching photo images of forensic department and the computer generated. In the entire process of implementation, the human based image decomposition is applied on the facial photo and computerized facial sketch after which feature extraction is performed followed by facial component regional weighting and classification. The result is then obtained. Facial recognition system is developed to match composite sketches to facial photographs. Experimental results were derived for the recognition system using datasets CUHK, FERET, TUPIS achieves accuracies of 80%, 81.5%, 88.5% respectively.

In their research article [31], authors have presented a novel technique to extract two image feature descriptor from components used for sketches to photo matching. In this article authors have initially focused on cropping and aligning the face sketches and photos together. Later based on the built-in attributes of composite sketches, Scale Invariant Feature Transform (SIFT) features and Histogram of Gradient (HOG) characteristics were drawn out against the images. The final outcomes of the recommended approach applied using e-PRIP database exhibited better performance. This technique achieved a result with accuracy 70.1%.

In article [32], a new approach for image representation was presented. A face recognition model was developed. The model was composed of many advanced techniques. CNN cascade was used for face detection and CNN for generating face buildings. For face recognition, a new approach for image augmentation was proposed. The development of Deep Learning Based Attendance System involves several important stages. Obtaining training dataset, augmentation, preparing images and then training Deep Neural Networks was done. Finally integration into existing system was performed in-order to test the method. With the proposed method of augmentation high accuracy can be achieved, 95.02% in overall.

Authors, in their research article [33] have implemented a methodology that addresses Deep Hyper sphere embedding for face recognition. This work developed a unique deep learning path for the recognition of faces. According to the proposal, a rendering technique called an angular softmax was implemented that can render good geometric interpretation by constricting the learned characteristics to be selective on a hyper sphere multiform. This technique of connecting the softmax with hyper sphere manifold generated made a softmax a very efficient and a good technique for face recognition in terms of the accuracy. The accuracy of face recognition and verification are 72.73% and 85.56% respectively. These accuracies already surpass most of the existing methods.

In an article [34], authors have proposed a precise and active Face Recognition Method Based on Hash Coding System. According to the methodology proposed in the article, technique of coding using hash function method and the network of cascaded type are constructed and implemented for the purpose of two step face recognition model. In the first stage, low geometric features and high dimensional features of each of the input image are drawn out in accordance with the various systems used for extraction. In the second stage, the low-dimensional features obtained from the first stage undergo the process of quantization to derive the codes called as hash by making use of a piecewise function. After these two stages, later by the calculation of distance between hash codes, identification is accomplished. Contemplating this approach and then trying to compare it with Visual Geometry Group (VGG), the performance of each image of Hash-VGG was found to be improved by some milliseconds and the accuracy was increased by 0.7%.

Research article [35] proposed by Galea, C., &Farrugia, R used a transfer learning model based on the traditional approach for face recognition. A novel model of 3D system called 3D morphable model was passed down to incorporate fresh images and unnaturally enlarge the training dataset. Synthetic type of multiple sketches was actually applied for the phase of testing to boost outcome, and the combination of the recommended system with a trending algorithm is presented to later increase the systems achievement. On comparing with major methods, implemented framework reduced rate of error by an amount of 80.7% for the sketches of viewed type.

In their research paper [36], authors have presented an approach of the type modified deep learning Neural Network system for the purpose of recognition of the faces. Proposed technique actually makes use of use of a Convolution Neural Networks (CNN) as its foundation. In the proposed technique, they have used a dataset that improves the generalization power of CNNs. The augmented training sets were then formulated such that they contain extra images generated by applying relevant filter techniques. Selecting a subset of an image from a dataset, synthetic type of images are later generated by applying a substantial noise factor using Poisson or Gaussian noise functions. Once after that, as a further part, the noise images are sampled to the set of images trained to construct the training set of an augmented type with twice the amount of samples. Augmented fragments are later supplied into the Convolution Neural Network (CNN) for the process of accomplishing training. Above trained network later tested using testing network and the rate of recognition is traced out based on the total number of correct matches produced for the test datasets with respect to the considered image set trained. CNN with Poisson's noise using about 8 datasets achieves accuracy 99.6%

In research article [37], Wang, M., Wang, Z., & Li, J. proposed a system that combines l LBP and a model of D-CNN. Proposed method extracted the features of LBP that corresponds to the facial image supplied as an input to the network of CNN, and a CNN network is trained with the help of LBP feature sets, and later makes use of the network that is trained for the recognition of face. This helped in avoiding the limitations of the weakest stability of gray scale of CNN model and can recognize the CNN trained model more accurately. The accuracy achieved using ORL andYALE datasets was about 96.6%

In their research article [38] entitled "Research on face recognition method based on deep learning in natural environment" presents a system of time invariant facial recognition using a matching system based on graph. In this research article authors have successfully implemented a graph-based representation system for age invariant face recognition. A simple deterministic algorithm is then used to identify the face of an individual in the contained datasets which also exploits topology of the graphs that is used for the purpose of matching which is considered as the second stage of the overall system. Experimental results are extracted on the FGnet Dataset. The used technique achieves an accuracy of about 99.94%.

In an article [39], authors have proposed the implementation of an evolutionary system of algorithm for composite sketch matching using an approach of transfer learning system. Transfer learning approach is majorly used in the methodology which aids in employing various types of sketch images accessible to study matcher of the most relevant type. Proposed technique initially performs the operation of pre-processing in which all the images are resized to 192*224 pixels. Then in the second step, extraction of various features is later accomplished on the set of images that are pre-processed, as a part of feature extraction, two different types of extractors are used to derive features: HIM (Histogram of Image Moments) and HOG (Histogram of Oriented Gradient). The former consists of moments corresponding to images that caters details pertaining to the information representing the orientation, pixel intensity etc. Histogram of Oriented Gradient was used to accomplish the assignment of detecting pedestrian. Methodology developed was considered as an evolutionary algorithm. Proposed system was also ably successful in solving the issue pertaining to the computer generated sketch recognition system using a novel evolutionary algorithm. Since there are very less composite datasets, other set of artist drawn and images of digital sets were also used. Results in this paper produced the accuracy of 34% accuracy for rank 10 of hand drawn sketches and accuracy of 5% for ran1 of computer sketches.

Research article [40] entitled "Face Recognition in Real-world Surveillance Videos with Deep Learning Method". In this article, authors have presented a novel dataset which are constructed out of surveillance frames from the real time destination. Later, a CNN having labelled sets of data was elegantly improved. Proposed method consists of 2 parts. First part consists of a set of data formed by automatically gathering and also data labelling out of the video surveillance frames. In the next stage, the face recognition model of VGG type is elegantly improved. Later the network on improvement accomplished the efficiency of 92.1 % of recognition.

In an article [41], research paper concentrated on accomplishing the task of verification of face and the task of individual re-recognition are addressed. Both of the tasks under an unconditional atmosphere are complex. This is because, dataset meant for the purpose of testing usually contains characters that are absent in the datasets. Hence to escape this complexity, a representation of deep discrimination system is used for learning a model that can cover both untrained and trained representations. From the input data, latent features are extracted. Deep discriminative representation learning achieving accuracies of 99.07% and 94.2% on the datasets LFW sets and YTF sets respectively.

In their research article [42], authors have proposed a Face Recognition system using Dominant Rotated Local Binary Pattern (DRLBP) and Scale Invariant Feature Transform (SIFT) Feature Extraction which was an innovative approach for classifying the images of human face using ANN(Artificial Neural Network ). Proposed technique is implemented in stages: In the first stage, all the facial images under the consideration are pre-processed. In the second stage, pre-processed image features are using SIFT. In the third stage extracted features of SIFT is then combined with DRLBP for the achievement of better accuracy. The accuracy of the existing system is 48%/ whereas the proposed system achieves 75% accuracy.

In their research article [43] entitled "Deep Learning on Binary Patterns for Face Recognition", authors aims in designing a powerful system for face recognition in real time. After pre-processing, number of standard filters were applied. Binary patterns are extracted and were supplied as an input into perception of multilayer to carry out the classification of image sets. Implemented system was put under the test on various datasets with challenges such as pose variations, occlusions etc. Method proposed delivered a better rate of efficiency in the neighbourhood of 91%.

In a research article [44] entitled "Matching Software-Generated Sketches to Face Photos with a Very Deep CNN, Morphed Faces, and Transfer Learning ", authors have proposed a technique that aims at matching software generated sketches to ace photos using deep CNN, faces that are morphed and an approach of transfer learning. For the purpose of synthesizing both photographs and corresponding sketches, a model called "3D morphable" is applied. Apart from that, VOM-SGFS database is extended to consist of a greater number of subjects. [45-46] This presented article also anticipated to hook the problems by using the consecutive offerings : (i) Deep-CNN was made used to compute the similarities of a subject in a computer generated sketch by correlating it with facial photographs, later then training is performed by using an approach of transfer learning to a model pre-trained for face recognition of face-photo, (ii) A model called "3D morphable" was applied to generate photos and computer generated sketches for the purpose of the augmentation on the set of training data that is reachable and (iii) Standard "UoM-SGFS" datasets is expanded to accommodate double the amount of sets after which contains 1200 sets of sketches corresponding to 600 subject sets. Results then proved the retrieval rate having an efficiency of above 90% for Rank100.

**3. OUTCOMES OF THE REVIEWED PAPERS**

On carrying out an exhaustive literature review, we have identified the outcomes of the various articles in terms of the deep learning technique used, their limitations and the accuracy rate of the system for various datasets used in the article. Table1. Presents the overview of different deep learning techniques used and their corresponding limitations identified in each of the article. Table 2. Presents the datasets used for the experiment and the accuracy rate achieved for the corresponding data set.

**Table1:** Overview of different papers in terms of the deep learning technique used and its limitations.

| rticle | Technique Used | Limitations |
|---|---|---|
| [5] | Histogram equalization and Homomorphic filtering | Technique failed to prove its impact on the performance factor |
| [7] | Metric Based approach | Difficult when there is some intra-class variations. Eg: pose |
| [8] | Face Detection and Face Landmark Estimation using Deep Learning | Cannot be used along text and audio features. |
| [11] | Boosted Deep Belief Network Framework | Cannot handle pose variations, video data etc. |
| [13] | Filter Pairing | Difficult to handle huge amount of data |
| [14] | Gradient Decent Approach | Failed to work when higher amount of attributes corresponding to the face is applied into the learning system |
| [15] | Principal Component Analysis, Binary Hashing, Block-wise Histograms | Dimension of the resulting feature resulting feature increases with the number of stages. |
| [17] | Joint line Fine Tuning | The performance of the concatenation method was worse than deep networks |
| [18] | Regularization approach | System miserably failed to collect large data sets relevant to the variations of an image with respect to the weight of a person |
| [19] | Deep Learning with Multitask Learning | Difficult when the face image has too many occlusions. |
| [20] | Robust Face Recognition | False acceptance rates cannot be decreased completely. |
| [21] | Histogram equalization and Homomorphic filtering | Technique failed to prove its impact on the performance factor |

| Article | Technique | Limitation |
|---|---|---|
| [22] | VGG Face Networks, Lightened CNN | Performance is not as high as state-of-the-art method. |
| [23] | Multi-instance Deep Learning | Cannot Handle 3D images |
| [26] | Feature Learning approach using Deep Neural Network (DNN) | Failed to sort out the ambiguities between pairs of very close expressions. |
| [27] | Transfer Learning | Cannot learn more layers above the source model. |
| [33] | Angular Softmax | Model proved in efficient when compared with Google FaceNet that is trained above 200 millions of data sets |
| [34] | Hash Coding | Loss function generated is not very accurate. |
| [36] | Modified deep learning | Uses supervised learning with human-annotated data. |
| [37] | Softmax-regression | Cannot be used to handle large DBs |
| [38] | Deep Learning in Natural Environment | Cannot solve problems of the fully-sided face and the severe blockade. |
| [40] | Experiments on Deep Learning | Cannot achieve very high recognition rate |
| [42] | DRLBP and SIFT Feature Extraction | Technique cannot be implemented in video-based features. |
| [44] | Transfer Learning | Can only be applied for general recognition of images. |

**Table 2:** Accuracy rate of various techniques on different datasets

| Article | Dataset Type | Dataset Name | Accuracy |
|---|---|---|---|
| [6] | Unlabeled images | Wild and ImageNet | 70% enhancement compared to other implemented methods. |
| [7] | Labeled faces | Wild (LFW) | 95.2% |
| [9] | Composite Sketches | FACES, SketchCop | 73% |
| [10] | Photo-Photo | CK+, SAIT, SAIT2, Internet | 97% for CK+ and SAIT 95.5% for SAIT2 and 84.5% with Internet Datasets |
| [11] | Labelled Face Images | CK+, JAFFE | 72% |
| [12] | Multiview images | Multi-Pie Dataset | 60-70% |
| [13] | Photo samples | CHUK03, VIPeR dataset | Rank-1 identification ranges from 15.66% to 19.89%. |
| [14] | Facial Images | FDDB dataset | 3% enhancement compared to other techniques |
| [15] | Labeled images | Wild (LFW), FERET | 78% |
| [16] | Labeled Photo-Photo Images | CK+, TFD | Using TFD, performance increased by 4.8% |
| [17] | Image Sequences | CK+ and Oulu-CASIA databases | 97.25% |
| [18] | Face Images | WIT database | Rank-1 accuracy=26% Rank-10 accuracy=65% |
| [19] | Photo Samples | Multi-Pie dataset | 4-6% enhancement over the other techniques |
| [22] | Photo-Photo samples | AR face Database, CMU PIE Database, Extended Yale Dataset | 98.95% |
| [23] | Photo-Photo samples | IJB-A, Wild(LFW), YTF, CASIA-Web Face Dataset | 97.15% of Mean accuracy |
| [24] | Face Images | FERA-2015, BP4D, SEMAINE | Accuracy is highest while using FERA-2015 |
| [25] | Image Samples | CT dataset | 99.9% |
| [26] | Facial Images | BU-3DFE and Multi-PIE | 85.2% |
| [29] | Photo Samples | FR, YALE | 92.8% accuracy was achieved using YALE. |
| [30] | Photo-Sketch samples | CUHK, FERET, TUPIS | 80%, 81.5%, 88.5% for CUHK, FERET, TUPIS |

| | | | datasets. |
|---|---|---|---|
| [31] | Composite Sketches-Photos | E-PRIP | 70.1% |
| [33] | Face Images | LFW, YTF | 72.3% and 85.56% |
| [36] | Facial Images | AT&T face database | 99.6% |
| [38] | Facial Images | CASIA, Wild(LFW), FGNet | 99.94% accuracy is achieved using FGNet databases. |
| [42] | Facial Images | CT dataset | 75% |

## 4. CONCLUSION

This review paper summarizes various techniques of Deep Learning used for facial recognition system. On performing an exhaustive literature review, it is very clearly found out that substantial amount of work is already accomplished under photo-photo identification in facial recognition system using Deep Learning Techniques. Many of the research articles have even proposed and implemented good number of works considering different variations like multi-expressions, time-invariant, weight variation, and illumination variation etc of photo-photo matching. After collecting various reviews, it is made very clear that a very few research articles have focused on the implementation of deep learning techniques for facial recognition system using forensic sketches to facial photograph matching. Hence still an admirable amount of scope for conducting an active research in the field of photo to sketch matching using Deep Learning techniques. On identifying various deep learning techniques used for facial recognition system, it is found out that very few papers have used transfer learning approach for the facial recognition system. So, in future a research could be concentrated in the area of photo to sketch matching using a deep learning technique with the combination of transfer learning approach which could prove a novel work. On accomplishment of the research in this area, researchers can possibly work out for boosting the performance of the system in identification. On inspection we have found many different data sets used for the purpose of the research and in future researchers have a very huge scope in building a dataset for photo-sketch images containing the photo-sketch sets of humans.